# Trajectory Smoothing Using GNSS/PDR Integration Via Factor Graph Optimization in Urban Canyons

Yihan Zhong, Weisong Wen*, and Li-Ta Hsu

*Abstract*—Smooth and accurate global navigation satellite system (GNSS) positioning for pedestrians in urban canyons is still a challenge due to the multipath effects and the non-light-of-sight (NLOS) receptions caused by the reflections from surrounding buildings. Factor graph optimization (FGO) attracts more and more attention in GNSS society for improving urban GNSS positioning by effectively exploiting the measurement redundancy from historical information to resist the outlier measurements. Unfortunately, the FGO-based GNSS standalone positioning is still challenged in highly urbanized areas. As an extension of the previous FGO-based GNSS positioning method, the potential of the pedestrian dead reckoning (PDR) model in FGO to improve the GNSS standalone positioning performance in urban canyons is exploited in this paper. Specifically, the relative motion of the pedestrian is estimated based on the raw acceleration measurements from the onboard smartphone inertial measurement unit (IMU) via the PDR algorithm. Then the raw GNSS pseudorange, Doppler measurements, and relative motion from PDR are integrated using the FGO. Given the context of pedestrian navigation with a small acceleration most of the time, a novel soft motion model is proposed to smooth the states involved in the factor graph model. This paper verified the effectiveness of employing the PDR model in FGO step-by-step through two datasets collected in dense urban canyons of Hong Kong using smartphone-level GNSS receivers. The comparison between the conventional extended Kalman filter, several existing methods, and FGO-based integration is presented. The results reveal that the existing FGO-based GNSS standalone positioning is highly complementary to the PDR's relative motion estimation. Both improved positioning accuracy and trajectory smoothness are obtained by utilizing the proposed GNSS/PDR integration method.

## I. Introduction

The urban canyon is one of the most economic scenarios, posing a significantly increased need for accurate pedestrian navigation [1, 2]. Nowadays, Global Navigation Satellite System (GNSS) receivers are still irreplaceable in the open area by providing reliable global positioning for users in urban.

Unfortunately, the GNSS performs poorly in urban areas like Hong Kong because of high-rise buildings. The obstacles would lead to the multipath effect and the non-light-of-sight (NLOS) [3]. The multipath effects and NLOS could potentially result in a significant error in GNSS positioning. Numerous researchers proposed methods to solve this problem by mitigating the impacts of the multipath effect and NLOS [4-7]. The 3D mapping aided the GNSS positioning method utilized the 3D city model to mitigate the impacts of the potential NLOS and multipath effects, such as the shadow matching algorithms [8, 9]. However, precise 3D building models are required for 3DMA GNSS [8, 9]. Moreover, the accuracy of the initial guess of the user's position profoundly affects their performance, which is difficult to obtain in deep urban canyons. Recently, factor graph optimization (FGO) based GNSS positioning methods [10] showed increased resistance against potential outliers, such as the NLOS and multipath. This is achieved by exploiting measurement redundancy from multiple historical epochs using the factor graph model, where mainly two consecutive epochs of information are exploited simultaneously in the conventional extended Kalman filtering (EKF) based method [11, 12]. A robust model was employed by the Chemnitz University of Technology team [13-15] to mitigate the outlier measurement effects in GNSS positioning using the FGO. In their work, only the raw GNSS pseudorange measurements were exploited. As an extension, we open-sourced a general FGO-based GNSS positioning framework [16], which exploited both the GNSS raw pseudorange and Doppler measurement. The result shows that the positioning performance was improved compared to the conventional EKF-based approach from the famous RTKLIB [17]. Interestingly, the recent FGO-based GNSS positioning developed by Dr. Suzuki [18] obtained the best accuracy in the Google Smartphone Challenged in ION GNSS+ 2021 [18] and ION GNSS+ 2022 [19]. His research used switchable constraints [20] together with batch optimization to maintain the trajectory smoothness. The above research showed that the FGO can mitigate the GNSS outlier's adverse effect by using the increased measurement redundancy from historical information. However, the FGO-based GNSS standalone positioning is still challenged in deep urban canyons, resulting in 20 meters of positioning error in highly urbanized areas [16].

Yihan Zhong, Weisong Wen and Li-Ta Hsu, are with the Department of Aeronautical and Aviation Engineering, Hong Kong Polytechnic University, Hong Kong (e-mail: yi-han.zhong@connect.polyu.hk; welson.wen@polyu.edu.hk; lt.hsu@polyu.edu.hk)

Inspired by the complementariness between the GNSS and the inertial measurement unit (IMU), the GNSS/IMU integration with the EKF was extensively investigated [21-25]. Similarly, the EKF estimator-based GNSS/ IMU integration can still be sensitive to the outlier measurements. As an extension, the FGO was also utilized to integrate GNSS/IMU navigation system [26-28]. [26] utilizing the IMU data redundancy to exclude faulty measurements. [27] investigated the Earth rotation impact regarding positioning accuracy and integrated the refined IMU pre-integration into an FGO-based navigation system to compensate for Earth rotation. [28] proposes an auto regressive integrated moving average (ARIMA) auxiliary model to overcome the GNSS signals interrupt in urban canyons and combine ARIMA and FGO to perform well even equipment outages occur. Although improved performance is obtained, those FGO-based GNSS/INS integration is still limited by the low measurement quality of the smartphone IMU during the occurrence of poor GNSS measurements as the INS-based propagation can quickly diverge without reliable correction from the GNSS. Different from the direct propagating of the raw IMU measurements, the pedestrian dead reckoning (PDR) [29-34] exploits the context of the pedestrian navigation which can provide periodical correction to the IMU by detecting the walking steps. A detailed review of PDR can be found at [35]. Inspired by this, researchers proposed to explore the potential of the GNSS/PDR integration [36-39] via the EKF estimator. For example, the [36] proposed an integrated GNSS/PDR method to narrow the gaps in the GNSS-denied situation and utilized peak detection, slide-window, and zero crossing methods to detect pedestrian steps. In [37], a real-time Kalman Filter-based GNSS/PDR position method was proposed with a multi-rate filter to prevent GNSS information from being unavailable. The [38] utilizes an EKF-based GNSS/PDR integration to improve the positioning performance of the smartphone-based navigation system in an environment that suffers the multipath effect. However, these EKF-based GNSS/PDR integration fails to explore the measurement redundancy from the historical information, which inspires us to raise a new question: *how would the PDR help with the existing FGO-based GNSS positioning as shown in Fig. 1*? Interestingly, the work in [39] investigated a loosely coupled GNSS/PDR integration using the FGO where PDR directly integrated with the position from the GNSS. Unfortunately, the loosely coupled integration in FGO could not fully exploit the strong time-correlation of the raw GNSS pseudorange measurements and the high-accuracy Doppler measurement was not exploited [12].

To effectively exploit the complementariness of the recently developed FGO-based GNSS positioning and the PDR, this paper investigates a tightly coupled FGO-based GNSS/PDR integration scheme for smartphone-based pedestrian navigation in urban canyons. On the one hand, the relative motion estimation of PDR can help resist the potential GNSS outliers. On the other hand, the GNSS raw measurements can help to correct the accumulated drift from PDR. More importantly, the global optimization from the FGO is exploited to relax the potential of both positioning sources to obtain an accurate and smooth trajectory which is of great significance for pedestrian navigation applications. Given the context of pedestrian navigation, this paper proposes a smoothness-driven motion model (SMM) which can effectively capture the dynamics of the pedestrian. The main contributions of this paper are summarized as follows:

(1) As an extension of our previous work in [16, 40], a tightly coupled FGO-based GNSS/PDR integrated positioning method is proposed in this paper. Specifically, Doppler and pseudorange measurements are tightly-coupled (TC) to connect the state set of two consecutive epochs in the factor graph. The relative motion is derived from the PDR based on the raw measurements from IMU which helps to establish an additional relative connection between the consecutive states.

(2) To effectively capture the dynamics of pedestrian navigation, this paper proposes a smoothness-driven motion model based on the context of pedestrian navigation with a small acceleration for trajectory smoothing which could also help to resist the potential GNSS outliers in GNSS standalone positioning.

(3) In this paper, the SMM and PDR factors' validity are progressively verified using a challenging dataset collected in the Hong Kong urban canyon with detaily discussion. Moreover, we compare the conventional EKF-based methods with several existing methods to our proposed GNSS/PDR integration method, again showing the proposed method's effectiveness.

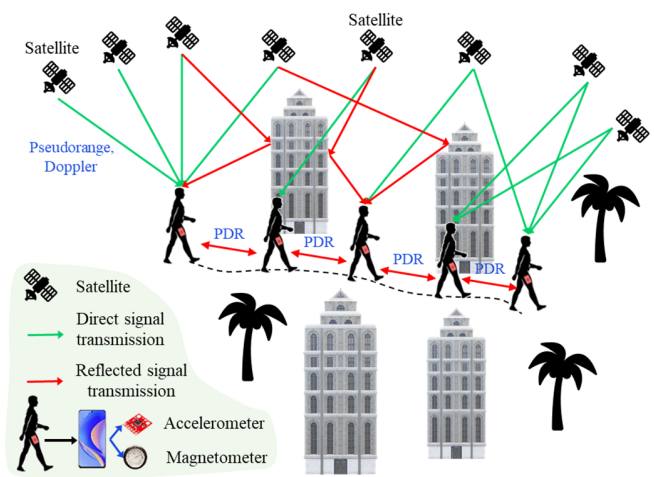

Fig. 1. Illustration of the challenges of the pedestrian navigation in urban canyons with numerous GNSS signal refelctions.

To the best of the author's knowledge, this work is the first research on the tightly-coupled integration of the Doppler and PDR measurements with FGO for a smoother trajectory of pedestrian navigation. The remainder of this paper is arranged as follows. Firstly, Section II overviews the proposed GNSS/PDR integration method. Then, the details of the PDR model and modeling of GNSS measurements were proposed in Section III. Furthermore, two experimental results in urban canyons will be interpreted in Section IV. section V presented discussions. Finally, Section VI presented conclusions and further work about the proposed GNSS/PDR integration.

## II. OVERVIEW OF THE PROPOSED METHOD

Our proposed FGO-based GNSS PDR positioning method is shown in Fig. 2. The input to the integrated GNSS/PDR system can be divided into two parts. The first part contains the raw pseudorange and Doppler measurements from the GNSS receiver in the smartphone. Satellites with elevation angles (ELE) below 15 degrees or signal-to-noise ratios (SNR) below 20 will be excluded. Another part includes the magnetometer and accelerometer measurements. The proposed GNSS/PDR integration output is the smartphone's state estimation. The GNSS/PDR positioning solution can be obtained by solving the formulated factor graph which contains the tightly-coupled Doppler factor, the pseudorange factor, the relative motion estimation from PDR, the smoothness-driven motion model, and the constant velocity (CV) factor in the optimization.

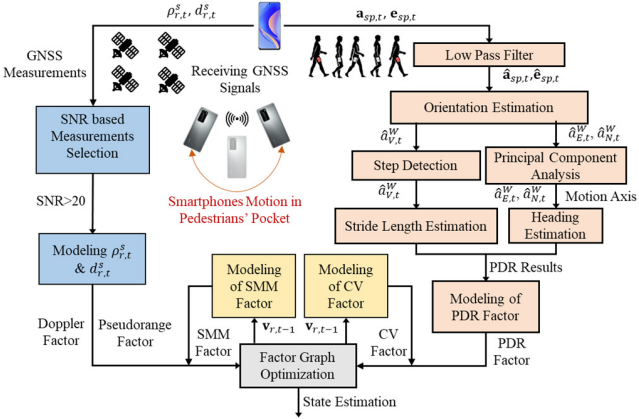

Fig. 2. Overview of the proposed FGO-based GNSS/PDR integration method.

Table 1. Symbols and their description in this paper

| Symbol | Description |
|---|---|
| $r$ | The GNSS receiver |
| $s$ | The index of the satellite |
| $\rho_{r,t}^s$ | The pseudorange measurement received from a satellite $s$ at a given epoch $t$ |
| $d_{r,t}^s$ | The Doppler measurement received from satellite $s$ at a given epoch $t$ |
| $\mathbf{p}_t^s$ | The position of the satellite $s$ at a given epoch $t$. $\mathbf{p}_t^s = (p_{t,x}^s, p_{t,y}^s, p_{t,z}^s)^T$ |
| $\mathbf{v}_t^s$ | The velocity of the satellite $s$ at a given epoch $t$. $\mathbf{v}_t^s = (v_{t,x}^s, v_{t,y}^s, v_{t,z}^s)^T$ |
| $\mathbf{p}_{r,t}$ | The position of the GNSS receiver at a given epoch $t$. $\mathbf{p}_{r,t} = (p_{r,t,x}, p_{r,t,y}, p_{r,t,z})^T$ |
| $\mathbf{v}_{r,t}$ | The velocity of the GNSS receiver at a given epoch $t$. $\mathbf{v}_{r,t} = (v_{r,t,x}, v_{r,t,y}, v_{r,t,z})^T$ |
| $\mathbf{a}_{sp,t}$ | The raw measurement from the smartphone accelerometers at a given epoch $t$. $\mathbf{a}_{sp,t} = (a_{sp,t,x}, a_{sp,t,y}, a_{sp,t,z})^T$ |
| $\mathbf{e}_{sp,t}$ | The raw measurement from the smartphone magnetometers at a given epoch $t$. $\mathbf{e}_{sp,t} = (e_{sp,t,x}, e_{sp,t,y}, e_{sp,t,z})^T$ |
| $\hat{\mathbf{a}}_{sp,t}$ | The filtered measurement from the accelerometers at a given epoch t in the global coordinate system. $\hat{\mathbf{a}}_{sp,t} = (\hat{a}_{sp,t,x}, \hat{a}_{sp,t,y}, \hat{a}_{sp,t,z})^T$ |
| $\hat{\mathbf{e}}_{sp,t}$ | The filtered measurement from the magnetometers at a given epoch t in the global coordinate system. $\hat{\mathbf{e}}_{sp,t} = (\hat{e}_{sp,t,x}, \hat{e}_{sp,t,y}, \hat{e}_{sp,t,z})^T$ |
| $\delta_{r,t}$ | The clock bias of the GNSS receiver at a given epoch $t$ |
| $\delta_{r,t}^s$ | The satellite clock bias at a given epoch $t$ |

To make the presentation of this paper clear, the notations in Table 1 are defined and are followed by the rest of this paper. Matrices are denoted in uppercase with bold letters. Vectors are denoted in lowercase with bold letters. Variable scalars are denoted as lowercase italic letters. Constant scalars are denoted as lowercase letters. Note that this paper is expressed the GNSS receiver's state, the output from PDR, and the position of satellites in the earth-centered, earth-fixed (ECEF) frame.

## III. TIGHTLY COUPLED GNSS/PDR POSITIONING VIA FGO

Fig.3 shows the factor graph of the proposed GNSS/PDR integrated positioning structure. The smartphone-level GNSS receiver state set is represented as follows:

$$\boldsymbol{\chi} = [\mathbf{x}_{r,1}, \mathbf{x}_{r,2}, \dots, \mathbf{x}_{r,t}, \dots, \mathbf{x}_{r,k}] \quad (1)$$

$$\mathbf{x}_{r,t} = (\mathbf{p}_{r,t}, \mathbf{v}_{r,t}, \delta_{r,t}, \dot{\delta}_{r,t})^T \quad (2)$$

where the variable $\boldsymbol{\chi}$ denotes the state set of the smartphone-level GNSS receiver ranging from the first epoch to the current epoch $k$. $\mathbf{x}_{r,t}$ denotes the single state of the smartphone at epoch $t$ which involves: the position ($\mathbf{p}_{r,t}$), velocity ($\mathbf{v}_{r,t}$), receiver clock bias ($\delta_{r,t}$) and clock drift ($\dot{\delta}_{r,t}$).

The CV and PDR factors are utilized to connect each state. To smooth the user's trajectories, the SMM factor is also included in our proposed factor graph, which is going to be elaborated in sub-section E. For a concise illustration, we briefly present the observation model and error function of pseudorange measurements which follow our previous work [16] but an explicit derivation of the Doppler factor which is tightly coupled in the system.

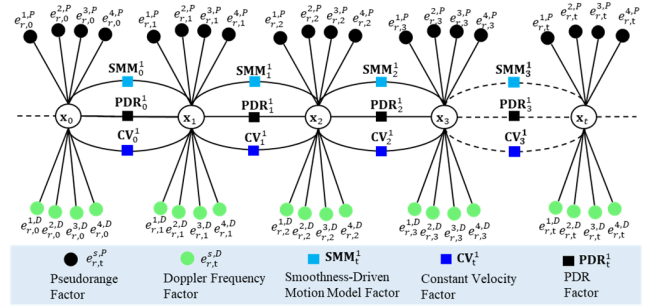

Fig. 3. The factor graph for proposed GNSS/PDR integration.

### A. Pseudorange Measurement Modeling

The observation model for GNSS pseudorange measurement from a given satellite $s$ is represented as follows:

$$\rho_{r,t}^s = h_{r,t}^s(\mathbf{p}_{r,t}, \mathbf{p}_t^s, \delta_{r,t}) + \omega_{r,t}^s \quad (3)$$

with $h_{r,t}^s(\mathbf{p}_{r,t}, \mathbf{p}_t^s, \delta_{r,t}) = ||\mathbf{p}_t^s - \mathbf{p}_{r,t}|| + \delta_{r,t}$

where the variable $\omega_{r,t}^s$ stands for the noise associated with the $\rho_{r,t}^s$. We can derive the error function ($\mathbf{e}_{r,t}^{s,P}$) in proposed GNSS/PDR integration for a given satellite measurement $\rho_{r,t}^s$ as follows:

$$||\mathbf{e}_{r,t}^{s,P}||_{\Sigma_{r,t}^s}^2 = ||\rho_{r,t}^s - h_{r,t}^s(\mathbf{p}_{r,t}, \mathbf{p}_t^s, \delta_{r,t})||_{\Sigma_{r,t}^s}^2 \quad (4)$$

where $\Sigma_{r,t}^s$ denotes the covariance matrix, calculated following the GO-GPS work [41] with the satellite SNR and ELE.

### B. Doppler Measurements Modeling

The range rate measurement vector ($\mathbf{y}_{r,t}^d$) at an epoch $t$ is expressed as follows:

$$\mathbf{y}_{r,t}^d = (\lambda d_{r,t}^1, \lambda d_{r,t}^2, \lambda d_{r,t}^3, \dots)^T \quad (5)$$

The $d_{r,t}^s$ and the $\lambda$ represent the Doppler measurement and the carrier wavelength of the satellite signal, respectively. The expected range rate $rr_{r,t}^s$ for satellite $s$ can also be calculated as follows:

$$rr_{r,t}^s = \mathbf{e}_{r,t}^{s,LOS}(\mathbf{v}_t^s - \mathbf{v}_{r,t}) + \frac{\omega_{earth}}{c_L}\left(v_{t,y}^s p_{r,t,x} + p_{t,y}^s v_{r,t,x} - p_{t,x}^s v_{r,t,y} - v_{t,x}^s p_{r,t,y}\right) + \dot{\delta}_{r,t} \quad (6)$$

where the variable $\omega_{earth}$ and $c_L$ denote the angular velocity of the earth's rotation [42] and the speed of light, respectively. The variable $\mathbf{e}_{r,t}^{s,LOS}$ denotes the line-of-sight vector connecting the GNSS receiver and the satellite. Furthermore, we can derive the error function ($\mathbf{e}_{r,k}^{s,D}$) of the tightly-coupled Doppler measurement as follows:

$$\|\mathbf{e}_{r,k}^{s,D}\|_{\Sigma_{r,t}^D}^2 = \|\lambda d_{r,t}^s - rr_{r,t}^s\|_{\Sigma_{r,t}^D}^2 \quad (7)$$

where $\Sigma_{r,t}^D$ denotes the covariance matrix corresponding to the Doppler measurement. And the $\Sigma_{r,t}^D$ is also calculated based on the ELE and SNR [41]. Due to the Doppler measurements being less sensitive to the GNSS multipath effects, $\Sigma_{r,t}^D$ is multiplied by a fixed coefficient valued at 10, which means larger weightings.

### C. Pedestrian Dead Reckoning Factor Modeling

#### 1) PDR Algorithm

The PDR factor is modeled by following the work in [40]. In the scenario where the pedestrian placed the smartphone in the pants pocket, the stability of the gyroscope is poor. Thus, only magnetometers and accelerometers are used. Both the magnetometer and accelerometer in the smartphone output 3D measurements in local coordinates defined by the smartphone, and the transformation of local coordinates to ECEF frame is necessary for global positioning. $\mathbf{a}_{r,t}$ and $\mathbf{e}_{r,t}$ are the accelerometer and the magnetometer measurements, respectively. Both of them are first used to determine the orientation of the device. And the measured acceleration could be converted to ECEF frame with the Android API *getRotationMatrix* [43], which aims to match the tri-axis acceleration and detected gravity to get the transformation to global coordinate. The three acceleration components are used to determine each step direction and step length. For attenuating the deterioration of dynamic pushes, a low-pass filtering is employed as shown below:

$$\hat{\mathbf{a}}_{sp,t} = (1 - \alpha^{acc})\mathbf{a}_{sp,t} + \alpha^{acc}(\hat{\mathbf{a}}_{sp,t-1}) \quad (8)$$

$$\hat{\mathbf{e}}_{sp,t} = (1 - \alpha^{mag})\mathbf{e}_{sp,t} + \alpha^{mag}(\hat{\mathbf{e}}_{sp,t-1}) \quad (9)$$

where the $\hat{\mathbf{a}}_{sp,i}$ denotes the filtered measurements of the accelerometer and $\hat{\mathbf{e}}_{sp,i}$ denotes the filtered measurements of the magnetometer. $\alpha^{acc}$ and $\alpha^{mag}$ denote the smoothing factor of the corresponding measurement field. Following the previous settings, we empirically set $\alpha^{acc}$ and $\alpha^{mag}$ as 0.6 and 0.84, respectively. The filtered measurements of acceleration can be calculated. The following part will concisely illustrate the detail of the PDR method.

(a) **The step detection**: Under normal pedestrian walking conditions, the vertical acceleration $\hat{a}_{V,t}^W$ reflects a distinct peak and depression for every foot of impact. This feature makes detecting the drop in vertical acceleration as the stride rate feasible. This paper empirically utilized a dip threshold at 7.5 m/s to ensure the correct detection.

(b) **The stride length estimation**: We followed the method proposed by Weinberg [44] based on the peak-to-peak of vertical acceleration to estimate the stride length, and the model is shown as:

$$l = K_w(a_{V,max} - a_{V,min})^{0.25} \quad (10)$$

where $l$ denotes the estimated stride length, $K_w$ denotes a constant for unit conversion which is empirically set as 0.713. $a_{V,max}$ and $a_{V,min}$ are the maximum and minimum vertical accelerations, respectively.

(c) **Heading direction estimation**: The employed PDR method estimated the heading direction in two steps. The first step is to perform a principal component analysis on the 2D plane with $\hat{a}_{E,t}^W$ and $\hat{a}_{N,t}^W$, excluding the vertical direction [30]. Then, they are subjected to the covariance matrix of Eigenvalue decomposition after smoothing the two sequences using a moving average. The largest eigenvalue result implies the direction estimation in the direction of parallel motion, which is the original forward direction. The second step of the heading direction estimation is to verify whether the original forward direction derived from the PCA is correct. According to [45], vertical and forward accelerations have a time relationship in the same cycle. Usually, the forward acceleration will be followed by a peak immediately after the step detection. By using this property, the forward direction could be decided.

#### 2) PDR Factor

The relative displacement of the smartphone $\Delta\mathbf{p}_{pdr,t}^{ecef}$ could be obtained after employing the PDR algorithm between two consecutive frames under the ECEF coordinate system:

$$\Delta\mathbf{p}_{pdr,t}^{ecef} = [\Delta x_t^{ecef}, \Delta y_t^{ecef}, \Delta z_t^{ecef}] \quad (11)$$

where $\Delta x_t^{ecef}$, $\Delta y_t^{ecef}$ and $\Delta z_t^{ecef}$ are displacement of the receiver which obtained from the PDR in three different directions. The difference between two consecutive epochs can be represented as:

$$\Delta\mathbf{p}_{r,t} = \mathbf{p}_{r,t} - \mathbf{p}_{r,t-1} \quad (12)$$

Where $\mathbf{p}_{r,t}$ denotes the position of the receiver at the *t* epoch and the $\mathbf{p}_{r,t-1}$ represents the position of the receiver at the *t-1* epoch. Hence, we can get the error function ($\mathbf{e}_{r,t}^R$) for a given relative displacement $\Delta\mathbf{p}_{pdr,t}^{ecef}$ as follows:

$$\|\mathbf{e}_{r,t}^R\|_{\Sigma_{r,t}^{PDR}}^2 = \|\Delta\mathbf{p}_{pdr,t}^{ecef} - \Delta\mathbf{p}_{r,t}\|_{\Sigma_{r,t}^{PDR}}^2 \quad (13)$$

where $\Sigma_{r,t}^{PDR}$ denotes the covariance matrix. We set it as the fixed value for denoting the performance of the PDR factor, which is experimentally determined as 0.1.

### D. Constant Velocity Factor

The constant velocity factor is constructed based on the assumption that the speed of the pedestrian is nearly invariant in walking. In addition to the velocity remaining constant during walking, the acceleration of the receiver should also converge to zero. Hence, we can get the observation model for the smartphone-level GNSS receiver velocity ($\mathbf{v}_{r,t}$) expressed as follows:

$$\mathbf{v}_{r,t} = h_{r,t}^C(\Delta \mathbf{p}_{r,t}, \mathbf{v}_{r,t}, \mathbf{v}_{r,t+1}) + \boldsymbol{\omega}_{r,t}^C \quad (14)$$

with $h_{r,t}^C(\Delta \mathbf{p}_{r,t}, \mathbf{v}_{r,t}, \mathbf{v}_{r,t+1}) =$
$$\begin{bmatrix} \Delta p_{r,t,x}/\Delta t - (v_{r,t,x} + v_{r,t+1,x})/2 \\ \Delta p_{r,t,y}/\Delta t - (v_{r,t,y} + v_{r,t+1,y})/2 \\ \Delta p_{r,t,z}/\Delta t - (v_{r,t,z} + v_{r,t+1,z})/2 \end{bmatrix},$$

where the $\mathbf{v}_{r,t}$ and $\mathbf{v}_{r,t+1}$ denote the velocity at two consecutive epochs with three-dimension in the ECEF frame, respectively. $\Delta p_{r,t,x}$, $\Delta p_{r,t,y}$, $\Delta p_{r,t,z}$ denote the receiver displacement between the $t$ frame and $t+1$ frame at three dimensions. The variable $\boldsymbol{\omega}_{r,t}^C$ denotes the noise associated with constant motion. Hence, the error function ($\mathbf{e}_{r,t}^C$) in our proposed GNSS/PDR integration for a given receiver velocity measurement $\mathbf{v}_{r,t}$ could be derived as follows:

$$||\mathbf{e}_{r,t}^C||_{\Sigma_{r,t}^C}^2 = ||h_{r,t}^C((\Delta \mathbf{p}_{r,t}, \mathbf{v}_{r,t}, \mathbf{v}_{r,t+1}))||_{\Sigma_{r,t}^C}^2 \quad (15)$$

where $\Sigma_{r,t}^C$ denotes the covariance matrix corresponding to the receiver velocity measurement. This paper empirically set $\Sigma_{r,t}^C$ as a fixed value.

*E. Smoothness-driven Motion Model Factor*

Given the context of pedestrian navigation, the accelerations caused by smartphone motion tend to be small. To exploit this property which can effectively improve the smoothness of the trajectory, this paper proposes a smoothness-driven motion model factor. We can calculate the acceleration of the smartphone-level GNSS receiver $\mathbf{a}_{r,t}$ as follows:

$$\mathbf{a}_{r,t} = \begin{bmatrix} (v_{r,t+1,y} - v_{r,t,y})/\Delta t \\ (v_{r,t+1,y} - v_{r,t,y})/\Delta t \\ (v_{r,t+1,z} - v_{r,t,z})/\Delta t \end{bmatrix} \quad (16)$$

Therefore, we can get the observation model for the acceleration ($\mathbf{a}_{r,t}$) expressed as follows:

$$\mathbf{a}_{r,t} = h_{r,t}^M(0, \mathbf{a}_{r,t}) + \boldsymbol{\omega}_{r,t}^M \quad (17)$$

with $h_{r,t}^M(0, \mathbf{a}_{r,t}) = \begin{bmatrix} 0 - (v_{r,t+1,y} - v_{r,t,y})/\Delta t \\ 0 - (v_{r,t+1,y} - v_{r,t,y})/\Delta t \\ 0 - (v_{r,t+1,z} - v_{r,t,z})/\Delta t \end{bmatrix}$

where the variable $\boldsymbol{\omega}_{r,t}^M$ denotes the noise associated with the zero acceleration model. Hence, we can get the error function ($\mathbf{e}_{r,t}^M$) in the proposed GNSS/PDR integration for a given receiver velocity measurement $\mathbf{v}_{r,t}$ as follows:

$$||\mathbf{e}_{r,t}^M||_{\Sigma_{r,t}^M}^2 = ||\mathbf{a}_{r,t} - h_{r,t}^M(0, \mathbf{a}_{r,t})||_{\Sigma_{r,t}^M}^2 \quad (18)$$

where $\Sigma_{r,t}^M$ denotes the covariance matrix corresponding to the receiver acceleration measurement. This paper also empirically set $\Sigma_{r,t}^M$ as a fixed value. Therefore, we can formulate the objective function for the GNSS/PDR integration using FGO based on the factors derived above as follows:

$$\boldsymbol{\chi}^* = \arg\min_{\boldsymbol{\chi}} \sum_{s,t} (||\mathbf{e}_{r,t}^D||_{\Sigma_{r,t}^D}^2 + ||\mathbf{e}_{r,t}^S||_{\Sigma_{r,t}^S}^2 + ||\mathbf{e}_{r,t}^R||_{\Sigma_{r,t}^{PDR}}^2 + ||\mathbf{e}_{r,t}^C||_{\Sigma_{r,t}^C}^2 + ||\mathbf{e}_{r,t}^M||_{\Sigma_{r,t}^M}^2) \quad (19)$$

the variable $\boldsymbol{\chi}^*$ denotes the optimal estimation of the state sets. The Ceres Solver [46] is used as a non-linear optimization solver for solving the objective function above. Meanwhile, the Levenberg-Marquardt (L-M) algorithm [47] is employed in our FGO processes to minimize the cost function iteratively.

IV. EXPERIMENT RESULTS AND DISCUSSION

*A. Experiment Setup*

We collected two datasets to verify the effectiveness of the proposed FGO-based GNSS/PDR positioning in two challenging scenarios (see Figures 4 and 5), which contain many buildings or trees, resulting in many potential multipath effects and NLOS reception in the scenes. During the tests, raw single-frequency GPS and BeiDou measurements were collected at 1 Hz using an Android-based Huawei P40 Pro phone [48]. The same smartphone recorded the accelerometer and magnetometer data via the Android API [48]. In the first experiment, the pedestrian's ground truth is labeled using a high-cost LiDAR/inertial positioning system based on the LIO-SAM [49]. Meanwhile, the loop-closure constraint is used to improve the LiDAR/inertial positioning system's accuracy, which can lead to sub-meter level ground truth positioning. In the second experiment, a highly densely urbanized area with dense traffic, we label the ground truth offline based on Google Earth and the timestamps shown to provide sub-meter accuracy by postprocessing carefully in our previous work [50]. We used an Intel i7-9750K 2.60GHz and a high-performance laptop with 32GB RAM to run the proposed framework.

*B. Evaluation Metrics and Methods*

Since the satellite geometry in dense urban canyons leads to very unreliable GNSS positioning in the vertical direction, in this paper, the horizontal positioning performance will be evaluated in the east, north, and up (ENU) framework, including the root-mean-square of error (RMSE), the mean error (MEAN), standard deviation (STD), maximum error (MAX).

To verify the effectiveness of our GNSS/PDR integration method in urban canyons step-by-step, we compared the following methods:

(1) **FGO**: GNSS pseudorange/Doppler integration using FGO [16].

(2) **EKF-PDR**: GNSS pseudorange/Doppler/PDR-based extended Kalman filter.

(3) **FGO-CV**: CV factor aided GNSS tightly-coupled Doppler/pseudorange integration using FGO.

(4) **FGO-CV-SMM** (1st proposed integration): The CV and SMM factors aided GNSS tightly-coupled Doppler/pseudorange integration using FGO.

(5) **FGO-PDR** (2nd proposed integration): The PDR factor aided GNSS tightly-coupled Doppler/pseudorange integration using FGO.

(6) **FGO-PDR-SMM** (3rd proposed integration): The PDR and SMM factors aided GNSS tightly-coupled Doppler/pseudorange integration using FGO.

(7) **FGO-PDR-CV** (4th proposed integration): The PDR and CV factors aided GNSS tightly-coupled Doppler/pseudorange integration using FGO.

(8) **FGO-PDR-CV--SMM** (5th proposed integration): The PDR, CV and SMM factors aided GNSS tightly-coupled Doppler/pseudorange integration using FGO.

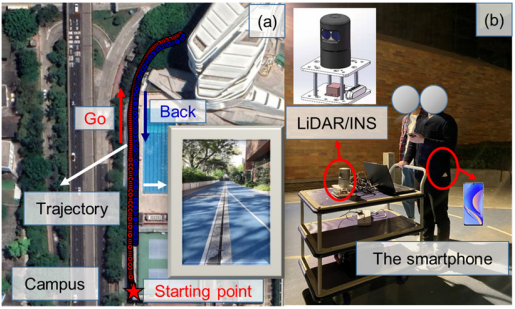

Fig. 4. The scenarios and sensor setup of the campus experiment. (a) the evaluated trajectory during the test. (b) the data collection platform with LiDAR/INS-based ground truth system and smartphone for GNSS data collection.

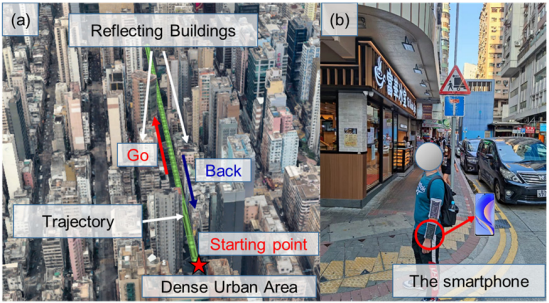

Fig. 5. The sensor setup of the dense urban area. (a) the evaluated trajectory in a dense urban canyon. (b) the data collection platform by hand-hold bag with a smartphone for GNSS data collection.

The colors corresponding to each method in the subsequent trajectory and error figures are as follows: The red denotes the FGO. The green color denotes the EKF-PDR. The light pink and pink denote FGO-CV and FGO-CV-SMM, respectively. The light orange and orange denote FGO-PDR and FGO-PDR-SMM, respectively. The light blue and blue colors denote FGO-PDR-CV, and FGO-PDR-CV-SMM, respectively.

### C. Experimental Evaluation on Campus

The 2D positioning trajectories in OpenStreetMap [51] and horizontal errors of all the listed positioning methods on the campus are shown in Figure 6 and Figure 7. From Figure 6, the trajectory of FGO-PDR-CV-SMM (blue curve) is consistent with the characteristics of pedestrian movement after applying the PDR factor, CV factor, and SMM factor. Figure 7 shows that adding the PDR factor could significantly reduce the positioning error around 50 and 150 epochs. Detailed statistic analyses of horizontal errors are shown in Table 2 and Table 3.

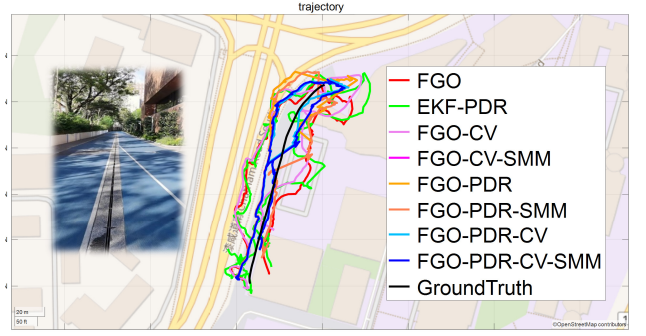

Fig. 6. Trajectories of evaluated eight methods on the campus in OpenStreetMap [51].

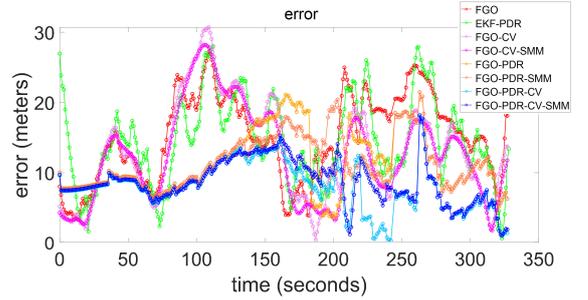

Fig. 7. Horizontal positioning errors on the campus.

Table 2. Positioning performance of the listed methods in the campus experiment

| All Methods | RMSE (m) | MEAN (m) | STD (m) | MAX (m) |
|---|---|---|---|---|
| FGO | 16.17 | 14.97 | 6.13 | 27.23 |
| EKF-PDR | 15.80 | 14.57 | 6.13 | 28.05 |
| FGO-CV | 14.85 | 13.13 | 6.94 | 30.69 |
| FGO-CV-SMM | 14.17 | 12.53 | 6.64 | 28.22 |
| FGO-PDR | 12.17 | 11.32 | 4.46 | 21.47 |
| FGO-PDR-SMM | 11.99 | 11.41 | 3.69 | 21.48 |
| FGO-PDR-CV | 8.63 | 7.97 | 3.33 | 17.98 |
| FGO-PDR-CV-SMM | 9.24 | 8.71 | 3.07 | 18.19 |

Table 3. Horizontal positioning improvements of the listed methods in the campus experiment

| Improvements | RMSE (m) | MEAN (m) | STD (m) | MAX (m) |
|---|---|---|---|---|
| FGO | / | / | / | / |
| EKF-PDR | 2.27% | 2.65% | 0.07% | -3.02% |
| FGO-CV | 8.19% | 12.26% | -13.10% | -12.74% |
| FGO-CV-SMM | 12.36% | 16.29% | -8.19% | -3.65% |
| FGO-PDR | 24.77% | 24.35% | 27.31% | 21.16% |
| FGO-PDR-SMM | 25.86% | 23.77% | 39.82% | 21.10% |
| FGO-PDR-CV | 46.61% | 46.77% | 45.69% | 33.95% |
| FGO-PDR-CV-SMM | 42.88% | 41.78% | 49.94% | 33.19% |

FGO can achieve 14.97 meters of mean error with 6.13 meters STD. However, judging from the fact that the maximum error of FGO still reaches 27.23 meters, the impact of unhealthy GNSS measurements on FGO is fatal. Applying the TC Doppler factor with the CV factor (FGO-CV) could decrease the mean error to 13.13 meters. This phenomenon shows that the CV factor with TC Doppler can effectively improve positioning performance. However, the CV factor based on pedestrian motion law lacks constraints on acceleration for resisting potential GNSS outliers which would lead to an unsmooth trajectory in GNSS-standalone positioning. After applying our proposed SMM factor, the mean error decreases to 12.53 meters which outperforms FGO-CV. However, the insufficiency in satellite number and unhealthy GNSS multipath effects would lead to limited improvement.

After applying the PDR to the TC Doppler FGO (FGO-PDR) instead of the CV factor, the mean error can achieve 11.32 meters with an STD of 4.46 meters. This improvement compared with the FGO-CV shows that the PDR factor could provide better relative position constraint between two consecutive epochs. Adding the SMM factor to FGO-PDR could achieve a lower mean of error. However, the improvement is still limited because the PDR factor contains the error source from the smartphone accelerometer and magnetometer which may not obey the pedestrian motion law. The effectiveness of the proposed CV factor could be verified with 7.97 meters mean error of FGO-PDR-CV, which adds the CV factor. Interestingly, the STD of error could achieve 3.07 meters after applying the SMM factor in FGO-CV-PDR-SMM, but with a larger mean of errors than FGO-CV-PDR. This is due to the occasional non-zero acceleration of the pedestrian during walking in this scenario.

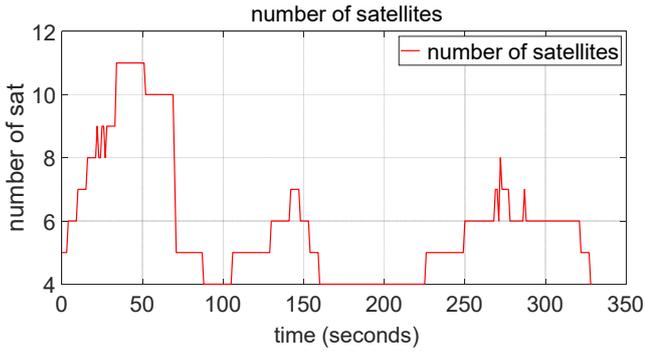

Fig. 8. The relationship between the number of satellites with time.

GNSS/PDR method based on EKF was also evaluated. Its RMSE is 15.80 m, slightly higher than FGO-CV and FGO-CV-SMM. This phenomenon indicates that our tightly-couple GNSS standalone FGO outperforms the EKF-PDR. The number of satellites in the campus experiment over time are shown in Figure 8. The number of satellites fluctuates during 75 to 150 epochs, which is not ideal for GNSS positioning. We can also see that the errors of FGO-PDR (light orange curve) are significantly smaller than the errors of FGO (red curve) and FGO-CV-SMM (pink curve) during this time which proves that our proposed PDR factor can still play a role in maintaining the need for pedestrian navigation in the GNSS denied environment.

### D. Experimental Evaluation in a Dense Urban Area

To challenge the effectiveness of the SMM factor and PDR factor for localization in complex scenes, we conducted another experiment in an even denser urban area. Table 4 and Table 5 show the above FGO-based GNSS/PDR integration methods positioning performances. The mean error of FGO is 16.80 in this scenario, which is larger than the one in the campus experiment (14.97 meters) due to the denser buildings which are more adverse for satellite signal transmission. After applying the SMM factor, the FGO-CV-SMM could outperform FGO. This phenomenon confirms that our proposed SMM factor can effectively resist GNSS outliers in dense urban areas. Interestingly, FGO-PDR and FGO-PDR-SMM show similar performance. FGO-PDR-CV decreased the error to 13.20 meters, which verified the effectiveness of the assumption of constant velocity in pedestrian navigation. After applying the SMM factor (FGO-PDR-CV-SMM), the mean error can decrease to 13.10 meters, which confirms the SMM factor's effectiveness in resisting potential GNSS outliers in GNSS/PDR integrating positioning in this scenario.

Table 4. Positioning performance of the listed methods in dense urban canyon

| All Methods | RMSE (m) | MEAN (m) | STD (m) | MAX (m) |
|---|---|---|---|---|
| FGO | 19.61 | 16.80 | 10.11 | 76.29 |
| EKF-PDR | 26.20 | 21.39 | 15.15 | 148.74 |
| FGO-CV | 20.53 | 17.50 | 10.74 | 108.94 |
| FGO-CV-SMM | 19.42 | 16.66 | 9.97 | 74.02 |
| FGO-PDR | 15.12 | 13.55 | 6.72 | 56.92 |
| FGO-PDR-SMM | 15.12 | 13.55 | 6.72 | 57.00 |
| FGO-PDR-CV | 14.51 | 13.20 | 6.04 | 50.01 |
| FGO-PDR-CV-SMM | 14.36 | 13.10 | 5.88 | 45.09 |

Table 5. Improvements of the listed methods in dense urban canyon

| Improvements | RMSE (m) | MEAN (m) | STD (m) | MAX (m) |
|---|---|---|---|---|
| FGO | / | / | / | / |
| EKF-PDR | -33.64% | -27.28% | -49.83% | -94.95% |
| FGO-CV | -4.72% | -4.16% | -6.25% | -42.80% |
| FGO-CV-SMM | 0.96% | 0.82% | 1.33% | 2.98% |
| FGO-PDR | 22.87% | 19.35% | 33.55% | 25.39% |
| FGO-PDR-SMM | 22.87% | 19.37% | 33.52% | 25.29% |
| FGO-PDR-CV | 25.98% | 21.46% | 40.22% | 34.45% |
| FGO-PDR-CV-SMM | 26.78% | 22.05% | 41.84% | 40.91% |

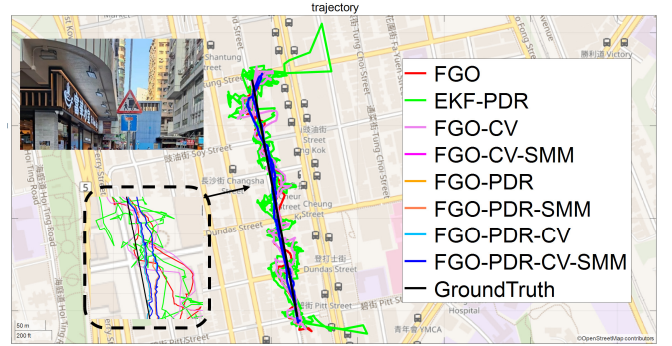

Fig. 9. Trajectories of evaluated eight methods in the urban canyon in OpenStreetMap [51].

Figures 9 and 10 show the 2D positioning trajectory in OpenStreetMap [51] and horizontal positioning errors in the dense urban scenario. In this scenario, the ground truth is a straight line, so we can easily compare the smoothness of all the methods. It is found that the GNSS pseudorange degrades due to signal reflections in dense urban environments would result in outliers that are terrible for the EKF-PDR (green curve). EKF-PDR results are worse than the GNSS-standalone FGO in this scenario, which shows a large error compared to the ground truth. Compared to FGO, FGO-PDR-CV-SMM shows a smoother trajectory.

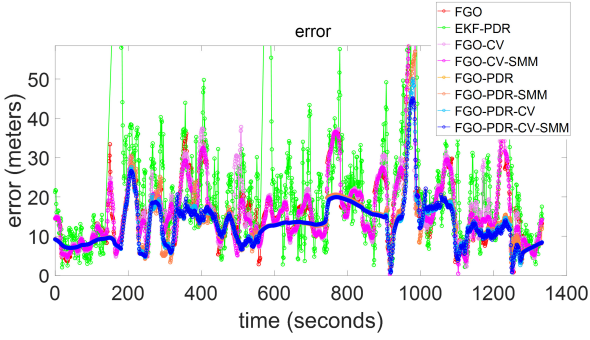

Fig. 10. Horizontal positioning errors in the dense urban experiment.

Figure 11 shows the number of satellites in the deep urban experiment with time. FGO-CV (light pink) and FGO-CV-SMM (pink curve), which without adding the PDR factor, reach an error close to 30 meters during epoch 250-400. The FGO-PDR (light orange curve) with the PDR factor added also shows resistance to the decreased data redundancy caused by the insufficient number of satellites.

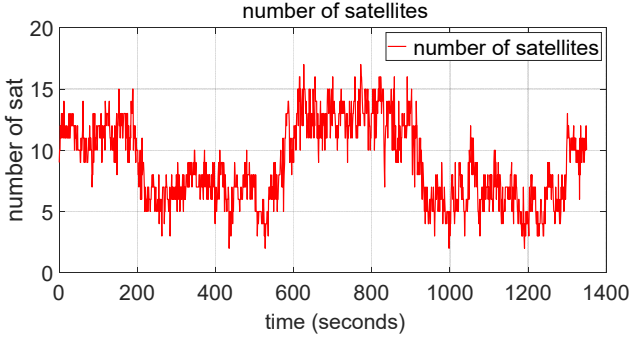

Fig. 11. The relationship between the number of satellites with time.

## V. DISCUSSIONS

### A. Discussion on PDR Results

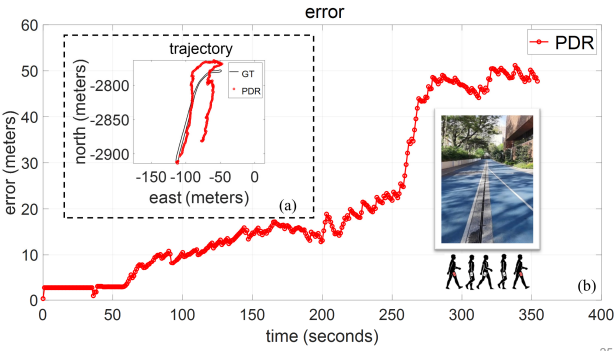

Fig. 12. (a) Horizontal PDR positioning trajectory in the campus experiment. The Ground Truth (GT) is the black curve. PDR trajectory is represented as the red curve. (b) 2D positioning error of the PDR in the campus experiment. The red curve denotes the PDR error.

Table 6 shows the performance of the employed PDR in the campus experiment. Figure 12 shows the employed PDR algorithm's horizontal trajectory and error. The PDR algorithm accumulated the errors over time, resulting in a very significant error in the PDR positioning results compared to the ground truth from the trajectory. The mean value of the positioning error of the PDR localization algorithm in the campus experiment is 21.04 m, while the RMSE is 26.74 m.

Table 6. Performance of the PDR in campus experiment

| Items | RMSE (m) | MEAN (m) | STD (m) | MAX (m) |
| --- | --- | --- | --- | --- |
| **PDR** | 26.74 | 21.04 | 16.52 | 51.18 |

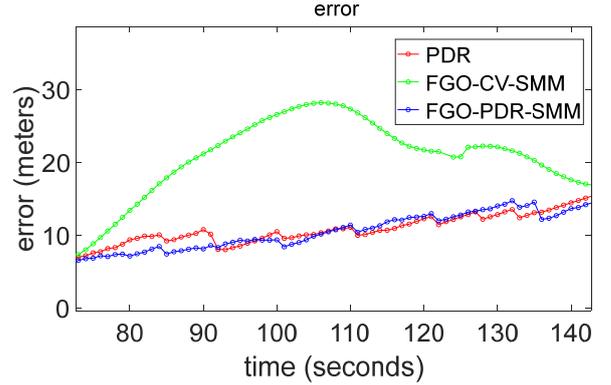

Fig. 13. 2D positioning error of PDR, FGO-CV-SMM, and FGO-PDR-SMM between 70 to 140 epochs in the campus experiment. The red, green, and cyan curves denote PDR, FGO-CV-SMM, and FGO-PDR-SMM, respectively.

Figure 13 shows the error between 70-140 seconds in the campus experiment. During this period, the PDR algorithm has lower errors than the FGO-CV-SMM, indicating that the proposed PDR factor provides better constraints than the CV factor in the estimation process.

### B. Discussion on GNSS Measurements Residuals

The performance improvement of FGO-PDR compared to FGO-CV-SMM demonstrate the effectiveness of the PDR factor. Regarding the two pipelines, the investigation of the residuals after the convergence from GNSS measurement models provides insights into the impact of the PDR factor on the derived solution. Ideally, the smaller residuals of applied measurements should potentially lead to more accurate estimation results, as a more consistent solution is derived given the assumption that the percentage of the healthy measurements exceeds the one from the polluted measurements. Figure 14 shows the histograms comparison of pseudorange and Doppler factor residuals after the optimization using the listed two methods. For the pseudorange measurement, it is shown that the peak of the FGO-CV-SMM pseudorange factor residuals around 0 is higher than the one from the FGO-PDR-CV-SMM. It is well-known that pseudorange is essential for the GNSS standalone positioning because it would offer the absolute position solution. The positioning results of FGO-PDR-CV-SMM can still perform better with larger pseudorange residuals, compared with the FGO-CV-SMM.

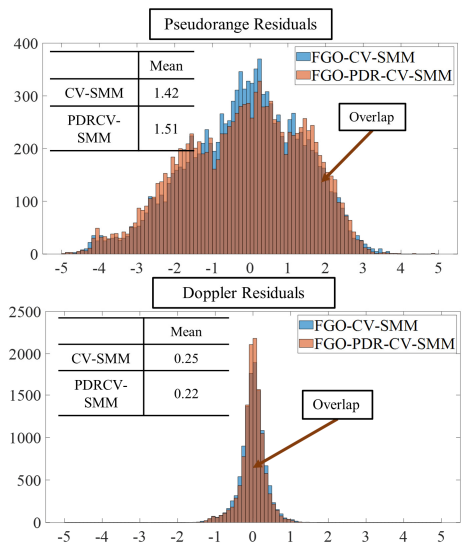

Fig. 14. Histograms of pseudorange factor and Doppler factor residuals. The x-axis denotes the value of residuals. The y-axis denotes the counts of residuals within the histogram.

Interestedly, a different phenomenon occurs in Doppler residuals. The mean value of the FGO-CV-SMM Doppler factor residuals is higher than the mean value of FGO-PDR-CV-SMM residuals. The potential reason is that the Doppler is less affected by the multipath effect while the pedestrian is walking along the street [52]. As a result, the assumption that the percentage of healthy measurements exceeds the one from polluted measurements could be largely held for the Doppler measurements.

## VI. CONCLUSIONS AND FUTURE WORK

Pedestrian navigation in urban canyons remains a challenging task. We propose an FGO-based GNSS/PDR positioning method in this paper, which tightly combines information from the GNSS raw measurements and the PDR derived from the magnetometer and accelerometer in the internal inertial measurement unit (IMU) of the smartphone for accurate pedestrian positioning and smoother trajectory estimation. We also verify the effectiveness of the assumption that pedestrians have small acceleration for trajectory smoothing. Applying the SMM factor could help resist the potential GNSS outliers. To effectively capture the dynamics of pedestrian navigation, this paper proposes a smoothness-driven motion model based on the GNSS/PDR localization results based on extended Kalman filtering are also evaluated in this paper, and the method exhibits significant degradation in dense urban canyons due to the influence of GNSS outliers in urban canyons.

We will employ carrier-phase measurements in our GNSS/PDR integration method in the future, providing more accurate positioning performance. We will also work on real-time sliding window-based FGO GNSS/PDR positioning to reduce the computational load for smartphone-level devices. We will also exploit collaborative positioning, which could improve measurement redundancy for better positioning performance in the future.

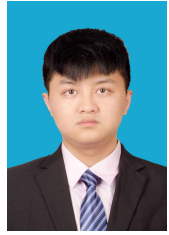

**Yihan Zhong** Yihan Zhong obtained his bachelor's degree in process equipment and control engineering from Guangxi University in 2020 and a Master's degree from The Hong Kong Polytechnic University (PolyU). He is currently a Ph.D. student at the Department of Aeronautical and Aviation Engineering (AAE) of PolyU.

His research interests include factor graph optimization-based collaborative positioning and low-cost localization.

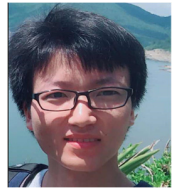

**Weisong Wen** was born in Ganzhou, Jiangxi, China. He received a Ph.D. degree in mechanical engineering, the Hong Kong Polytechnic University. He was a visiting student researcher at the University of California, Berkeley (UCB) in 2018. He is currently a research assistant professor in the Department of Aeronautical and Aviation Engineering.

His research interests include multi-sensor integrated localization for autonomous vehicles, SLAM, and GNSS positioning in urban canyons.

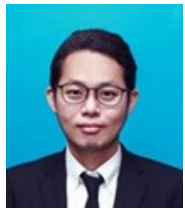

**Li-Ta Hsu** received the B.S. and Ph.D. degrees in aeronautics and astronautics from National Cheng Kung University, Taiwan, in 2007 and 2013, respectively. He is currently an assistant professor with the Department of Aeronautical and Aviation Engineering. The Hong Kong Polytechnic University, before he served as a post-doctoral researcher in the Institute of Industrial Science at the University of Tokyo, Japan. In 2012, he was a visiting scholar at University College London, the U.K.

His research interests include GNSS positioning in challenging environments and localization for pedestrian, autonomous driving vehicle, and unmanned aerial vehicle.